\documentclass[sigconf]{acmart}

\usepackage{CJKutf8}
\usepackage{natbib}
\usepackage{xspace}
\usepackage{subcaption}

\usepackage{color}

\usepackage[utf8]{inputenc}
\usepackage[T1]{fontenc}
\usepackage{hyphenat}
\usepackage{xspace}
\usepackage{amsmath}
\usepackage{amsfonts}
\usepackage{hyperref}
\usepackage{url}
\usepackage{booktabs}
\usepackage{multirow}
\usepackage{makecell}
\usepackage{caption}
\usepackage{minibox}
\usepackage{bbm}
\usepackage{graphicx}
\usepackage{balance}
\usepackage{mathtools}
\usepackage{color}
\usepackage{marvosym}
\usepackage{ifthen}
\usepackage{textcomp}
\usepackage{enumitem}
\usepackage{verbatim}
\usepackage{algorithm}
\usepackage{numprint}
\usepackage{balance}

\usepackage{amsthm}
\theoremstyle{plain}

\newcommand{\chatoDisplayMode}[1]{#1}

\definecolor{MyRed}{rgb}{0.6,0.0,0.0} 
\definecolor{MyBlack}{rgb}{0.1,0.1,0.1} 
\newcommand{\inred}[1]{{\color{MyRed}\sf\textbf{\textsc{#1}}}}
\newcommand{\frameit}[2]{
  \begin{center}
  {\color{MyRed}
  \framebox[.9\columnwidth][l]{
    \begin{minipage}{.85\columnwidth}
    \inred{#1}: {\sf\color{MyBlack}#2}
    \end{minipage}
  }\\
  }
  \end{center}
}

\newcommand{\note}[2][]{\chatoDisplayMode{\def\@tmpsig{#1}\frameit{{\Pointinghand} Note}{#2\ifx \@tmpsig \@empty \else \mbox{ --\em #1}\fi}}}
\newcommand{\todo}[2][]{\chatoDisplayMode{\def\@tmpsig{#1}\frameit{{\Writinghand} To-do}{#2\ifx \@tmpsig \@empty \else \mbox{ --\em #1}\fi}}}

\newcommand{\abbrevStyle}[1]{#1}

\newcommand{\ie}{\abbrevStyle{i.e.}\xspace}
\newcommand{\eg}{\abbrevStyle{e.g.}\xspace}
\newcommand{\cf}{\abbrevStyle{cf.}\xspace}

\newcommand{\vs}{\abbrevStyle{vs.}\xspace}
\newcommand{\etc}{\abbrevStyle{etc.}\xspace}

\newcommand{\Secref}[1]{Sec.~\ref{#1}}

\newcommand{\Tabref}[1]{Table~\ref{#1}}
\newcommand{\Figref}[1]{Fig.~\ref{#1}}

\newcommand{\Appref}[1]{Appendix~\ref{#1}}

\newcommand{\xhdr}[1]{\vspace{1.7mm}\noindent{{\bf #1.}}}

\newcommand{\textcite}[1]{\citeauthor{#1} \shortcite{#1}}

\newcommand{\cpt}[1]{\textsc{\MakeLowercase{#1}}}

\newcommand{\hide}[1]{}

\makeatletter
\newcommand{\iffont}[2]{\ifthenelse{\equal{\f@family}{#1}}{#2}{}}
\makeatother

\iffont{ptm}{
  \usepackage{mathptmx}

  \DeclareSymbolFont{greek}{OML}{cmm}{m}{n}
  \DeclareMathSymbol{\alpha}{\mathalpha}{greek}{"0B}
  \DeclareMathSymbol{\beta}{\mathalpha}{greek}{"0C}
  \DeclareMathSymbol{\gamma}{\mathalpha}{greek}{"0D}
  \DeclareMathSymbol{\delta}{\mathalpha}{greek}{"0E}
  \DeclareMathSymbol{\epsilon}{\mathalpha}{greek}{"0F}
  \DeclareMathSymbol{\zeta}{\mathalpha}{greek}{"10}
  \DeclareMathSymbol{\eta}{\mathalpha}{greek}{"11}
  \DeclareMathSymbol{\theta}{\mathalpha}{greek}{"12}
  \DeclareMathSymbol{\iota}{\mathalpha}{greek}{"13}
  \DeclareMathSymbol{\kappa}{\mathalpha}{greek}{"14}
  \DeclareMathSymbol{\lambda}{\mathalpha}{greek}{"15}
  \DeclareMathSymbol{\mu}{\mathalpha}{greek}{"16}
  \DeclareMathSymbol{\nu}{\mathalpha}{greek}{"17}
  \DeclareMathSymbol{\xi}{\mathalpha}{greek}{"18}
  \DeclareMathSymbol{\pi}{\mathalpha}{greek}{"19}
  \DeclareMathSymbol{\rho}{\mathalpha}{greek}{"1A}
  \DeclareMathSymbol{\sigma}{\mathalpha}{greek}{"1B}
  \DeclareMathSymbol{\tau}{\mathalpha}{greek}{"1C}
  \DeclareMathSymbol{\upsilon}{\mathalpha}{greek}{"1D}
  \DeclareMathSymbol{\phi}{\mathalpha}{greek}{"1E}
  \DeclareMathSymbol{\chi}{\mathalpha}{greek}{"1F}
  \DeclareMathSymbol{\psi}{\mathalpha}{greek}{"20}
  \DeclareMathSymbol{\omega}{\mathalpha}{greek}{"21}
  \DeclareMathSymbol{\varepsilon}{\mathalpha}{greek}{"22}
  \DeclareMathSymbol{\vartheta}{\mathalpha}{greek}{"23}
  \DeclareMathSymbol{\varpi}{\mathalpha}{greek}{"24}
  \DeclareMathSymbol{\varrho}{\mathalpha}{greek}{"25}
  \DeclareMathSymbol{\varsigma}{\mathalpha}{greek}{"26}
  \DeclareMathSymbol{\varphi}{\mathalpha}{greek}{"27}
  \DeclareSymbolFont{otone}{OT1}{cmr}{m}{n}
  \DeclareMathSymbol{\Gamma}{\mathalpha}{otone}{0}
  \DeclareMathSymbol{\Delta}{\mathalpha}{otone}{1}
  \DeclareMathSymbol{\Theta}{\mathalpha}{otone}{2}
  \DeclareMathSymbol{\Lambda}{\mathalpha}{otone}{3}
  \DeclareMathSymbol{\Xi}{\mathalpha}{otone}{4}
  \DeclareMathSymbol{\Pi}{\mathalpha}{otone}{5}
  \DeclareMathSymbol{\Sigma}{\mathalpha}{otone}{6}
  \DeclareMathSymbol{\Upsilon}{\mathalpha}{otone}{7}
  \DeclareMathSymbol{\Phi}{\mathalpha}{otone}{8}
  \DeclareMathSymbol{\Psi}{\mathalpha}{otone}{9}
  \DeclareMathSymbol{\Omega}{\mathalpha}{otone}{10}
  \DeclareSymbolFont{syms}{OML}{cmm}{m}{it}
  \DeclareMathSymbol{\partial}{\mathord}{syms}{"40}
  \DeclareMathAlphabet{\mathbold}{OML}{cmm}{b}{it}
  \DeclareSymbolFont{largesymbols}{OMX}{cmex}{m}{n}

}

\AtBeginDocument{%
  \providecommand\BibTeX{{%
    \normalfont B\kern-0.5em{\scshape i\kern-0.25em b}\kern-0.8em\TeX}}}

\copyrightyear{2023} 
\acmYear{2023} 
\setcopyright{acmlicensed}\acmConference[WWW '23]{Proceedings of the ACM Web Conference 2023}{April 30--May 4, 2023}{Austin, TX, USA}
\acmBooktitle{Proceedings of the ACM Web Conference 2023 (WWW '23), April 30--May 4, 2023, Austin, TX, USA}
\acmPrice{15.00}
\acmDOI{10.1145/3543507.3583220}
\acmISBN{978-1-4503-9416-1/23/04}

\begin{document}

\title{Descartes: Generating Short Descriptions of Wikipedia Articles}


\author{Marija \v{S}akota}
 \affiliation{%
   \institution{EPFL}
   \country{Switzerland}}
\email{marija.sakota@epfl.ch}

\author{Maxime Peyrard}
 \affiliation{%
   \institution{EPFL}
   \country{Switzerland}}
\email{maxime.peyrard@epfl.ch}

\author{Robert West}
 \affiliation{%
   \institution{EPFL}
   \country{Switzerland}}
\email{robert.west@epfl.ch}

\renewcommand{\shortauthors}{\v{S}akota, Peyrard, and West}

\newcommand\maxime[1]{\textcolor{teal}{[Maxime: #1]}}

\hyphenation{
Wi-ki-pe-dia
Wi-ki-me-dia
Wi-ki-da-ta
De-ter-mine
Page-Rank
web-page
web-pages
da-ta-set
}

\newcommand{\name}{Descartes\xspace}
\newcommand{\WP}{Wikipedia\xspace}
\newcommand{\sd}{short description\xspace}
\newcommand{\Sd}{Short description\xspace}
\newcommand{\lang}{l}
\newcommand{\tgtLang}{L}
\newcommand{\desc}{description\xspace}
\newcommand{\descs}{descriptions\xspace}
\newcommand{\descText}[1]{\textit{``#1''}}

\begin{abstract}
Wikipedia is one of the richest knowledge sources on the Web today.
In order to facilitate navigating, searching, and maintaining its content, Wikipedia's guidelines state that all articles should be annotated with a so-called short description indicating the article's topic
(\eg, the short description of \cpt{Beer} is \descText{Alcoholic drink made from fermented cereal grains}).
Nonetheless, a large fraction of articles (ranging from 10.2\% in Dutch to 99.7\% in Kazakh) have no short description yet, with detrimental effects for millions of Wikipedia users.
Motivated by this problem, we introduce the novel task of automatically generating short descriptions for Wikipedia articles and propose Descartes, a multilingual model for tackling it.
Descartes integrates three sources of information to generate an article description in a target language:
the text of the article in all its language versions,
the already-existing descriptions (if any) of the article in other languages,
and semantic type information obtained from a knowledge graph.
We evaluate a Descartes model trained for handling 25 languages simultaneously, showing that it beats baselines (including a strong translation-based baseline) and performs on par with monolingual models tailored for specific languages.
A human evaluation on three languages further shows that the quality of Descartes's descriptions is largely indistinguishable from that of human-written descriptions; \eg, 91.3\% of our English descriptions (\vs\ 92.1\% of human-written descriptions) pass the bar for inclusion in Wikipedia, suggesting that Descartes is ready for production, with the potential to support human editors in filling a major gap in today's Wikipedia across languages.

\end{abstract}





\maketitle

\section{Introduction}
\label{sec:introduction}

\begin{figure}[t]
    \centering
    \begin{subfigure}{0.32\columnwidth}
    \includegraphics[width=\textwidth]{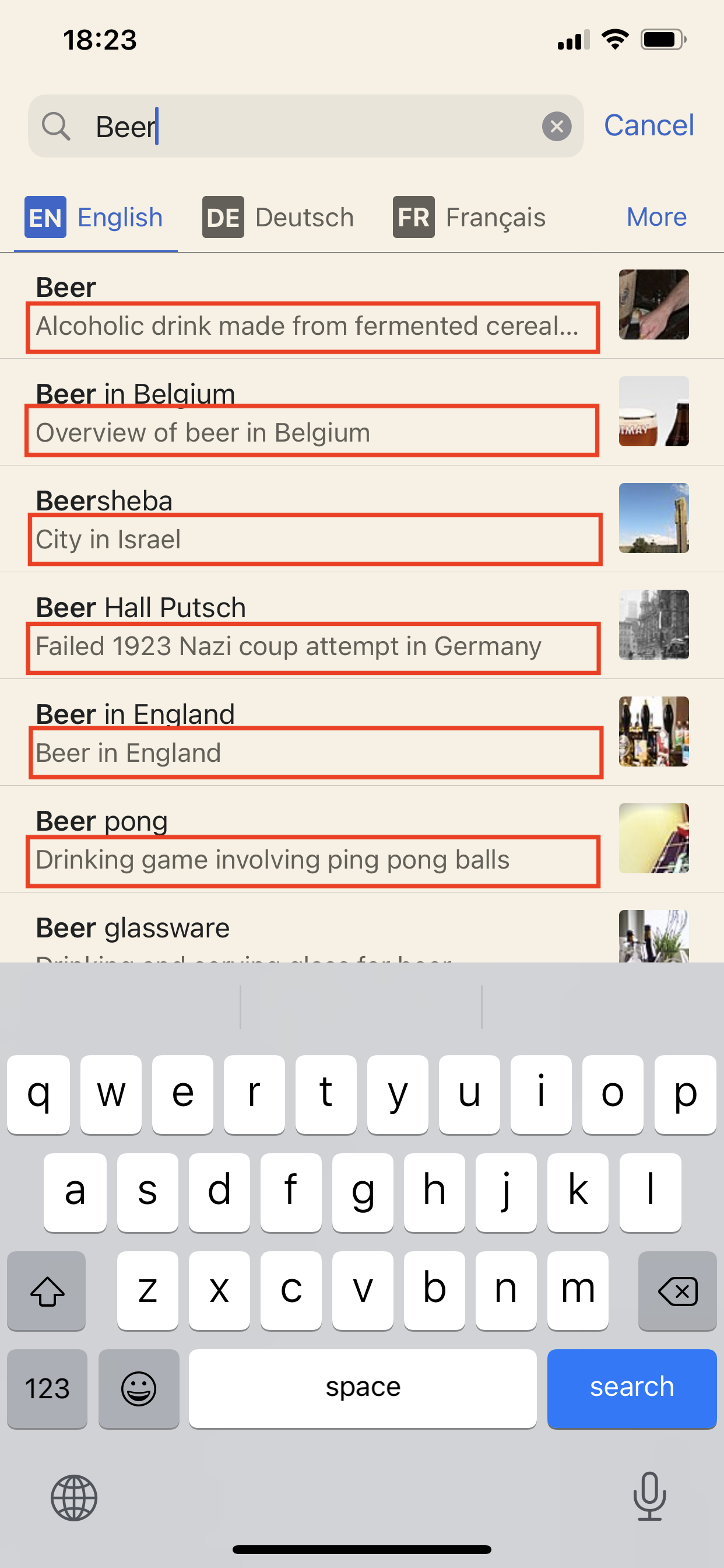}
    \caption{Search}
    \label{fig:screenshots_a}
    \end{subfigure}
    \begin{subfigure}{0.32\columnwidth}
    \includegraphics[width=\textwidth]{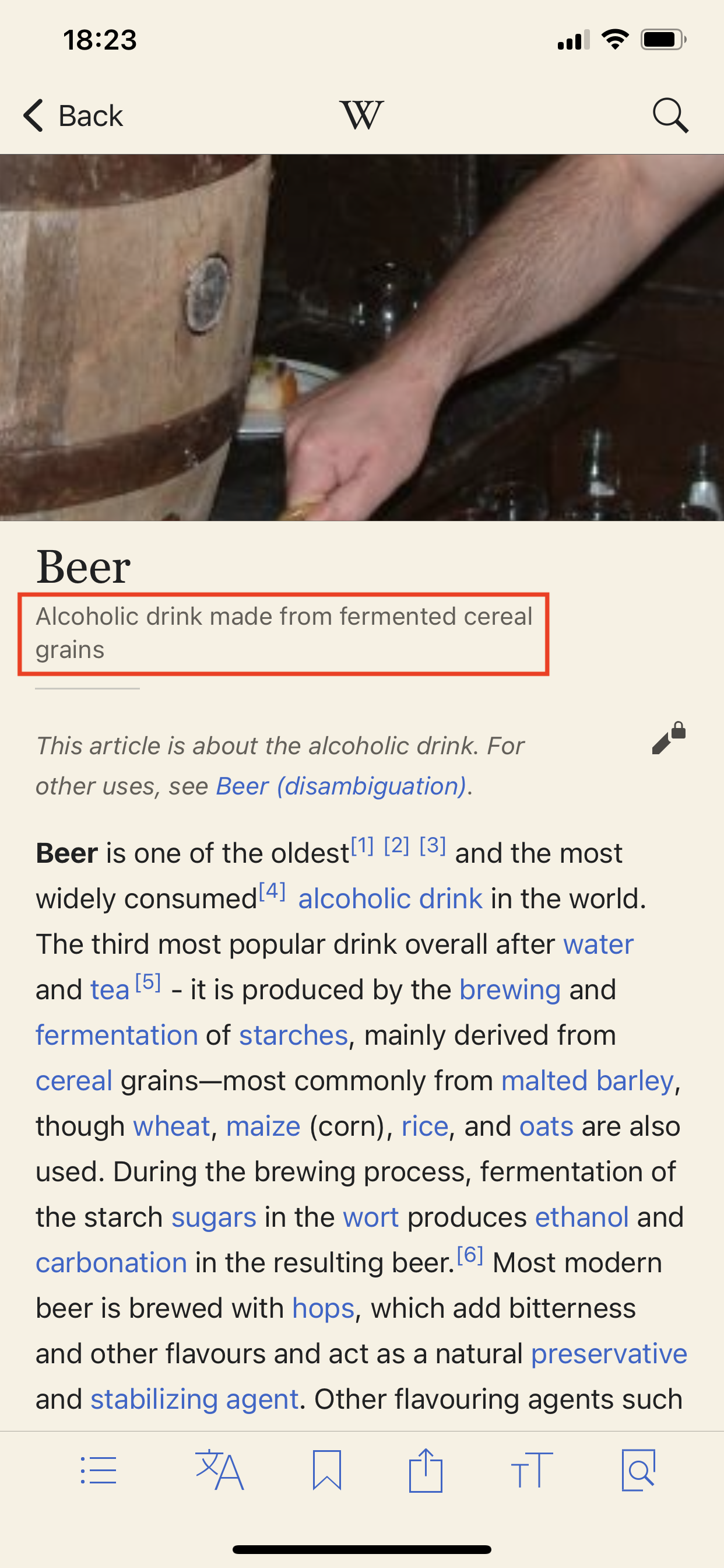}
    \caption{Summary}
    \label{fig:screenshots_b}
    \end{subfigure}
    \begin{subfigure}{0.32\columnwidth}
    \includegraphics[width=\textwidth]{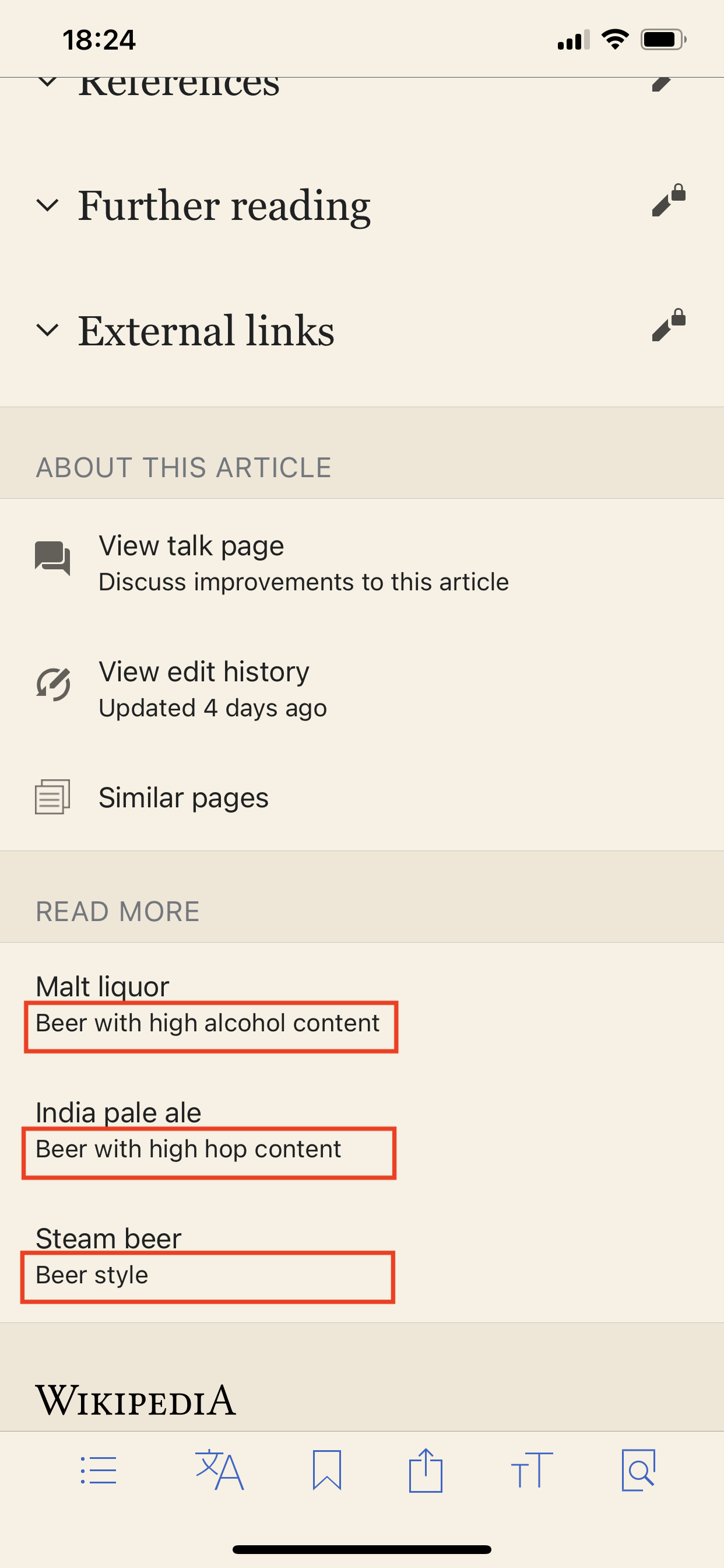}
    \caption{Read more}
    \label{fig:screenshots_c}
    \end{subfigure}
    \vspace{-2mm}
    \caption{Three use cases of \sd{}s on \WP.}
    \label{fig:screenshots}
    \vspace{-2mm}
\end{figure}



With over 42M articles in around 300 languages, \WP is the largest encyclopedia ever built.
Since most people are unfamiliar with most entities covered by \WP articles, \WP{}'s guidelines stipulate that each article should be annotated by a so-called \emph{\sd{}} providing
``a very brief indication of the field covered by the article''.%
\footnote{
\label{fn:SD instructions}
See \url{https://en.wikipedia.org/wiki/Wikipedia:Short_description};
regarding the relationship with Wikidata's item descriptions, see \Appref{appendix:wikidata}.
}
For instance, the \sd of \cpt{Beer} is \descText{Alcoholic drink made from fermented cereal grains}.


When shown together with article titles, \sd{}s can vastly facilitate navigation and search, as shown in \Figref{fig:screenshots}, which exemplifies how \sd{}s are used in the \WP mobile app to
(a)~disambiguate search results,
(b)~summarize an article's topic at the top of the page, and
(c)~annotate links in the ``read more'' section at the bottom of the page.
Beyond knowledge consumption, \sd{}s are also useful for knowledge production and management;
\eg, editors rely heavily on lists of article titles for organizing their work, and annotating these titles with \sd{}s can increase intuition and efficiency.


Although \WP{}'s guidelines require a \sd for each article, a large fraction of all articles do not have one, as shown in \Tabref{tab:wp-stats} for 25 language editions.
The problem is particularly striking for low-resource languages (\eg, 99.7\% of Kazakh and 95.1\% of Lithuanian articles have no short description), but it also strongly affects high-resource languages (\eg, 19.7\% of English and 19.1\% of German articles have no \sd).
Across the 25 languages of \Tabref{tab:wp-stats},
9.3M articles (43\%) have no \sd, with detrimental effects on the navigability, retrievability, and maintainability of \WP's content for hundreds of millions of users.


As volunteer time is notoriously scarce on \WP---there is vastly more work that should than that could be done by human editors \cite{vrandecic_building_2021}---writing the millions of missing \sd{}s is not feasible with human labor alone, but requires automated tools.
Building these tools is the focus of this paper.


The task of generating \sd{}s for \WP articles occupies a sweet spot, being (1)~\emph{within reach} yet (2)~\emph{challenging,} with the potential for (3)~\emph{high-impact} applications.

First, the task is \emph{within reach} as the information required to solve it is usually present somewhere in the respective article.

Second, the task remains \emph{challenging} as standard methods do not suffice:
Hand-coded rules (\eg, regular expressions \cite{hearst-1992-automatic}) are insufficient given the extreme heterogeneity of the mapping from article texts to \sd{}s;
in many cases the ideal \desc does not appear explicitly as a substring of the article text at all, but needs to be combined from multiple sentences (\cf\ example of \Figref{fig:screenshots_b});
and the concept in question may not yet have an article in the target language to begin with from which a \sd could be extracted by a rule.
Machine translation, which could be used to translate already\hyp existing descriptions of the concept in question from other languages into the target language, is also insufficient, since it is not clear from which of the potentially many languages to translate; ideally, one would like to pool information from the already\hyp existing descriptions across all languages, combined with additional textual modalities such as article texts and non-textual modalities such as knowledge graphs.
For these reasons, the task also goes beyond conventional text summarization (\cf\ \Secref{sec:background}).

Finally, tackling the task has the potential for \emph{high impact} as \sd{}s are such an important, yet overwhelmingly incomplete, feature of \WP, while the threshold for incorporating auto\hyp generated \descs in live \WP is low, compared to other kinds of auto\hyp generated content such as entire article texts \cite{sauper-barzilay-2009-automatically, liu2018generating, banerjee-mitra-2015-wikikreator, fan-gardent-2022-generating}, infoboxes \cite{saez2018, lange2010, yus2014}, \etc:
\WP requires all edits to be vetted by humans, which is feasible with relatively little training (and on small, intuitive user interfaces) for \sd{}s (\cf\ footnote~\ref{fn:SD instructions}), whereas it requires experienced editors (and more complex user interfaces) for other types of content such as entire articles or infoboxes.
Incorporating auto\hyp generated \descs is thus well suited for helping recruit and onboard new editors, instead of further straining the already\hyp overloaded existing editors.


\xhdr{Proposed solution}%
\footnote{
\label{fn:code and data}
Code, data, and models available at \url{https://github.com/epfl-dlab/descartes}}
We tackle the task with \emph{\name} (short for ``\emph{Desc}riber of \emph{art}icl\emph{es}''), a generative language model that integrates three modalities to produce \sd{}s of a \WP article about entity $E$ in language $\tgtLang$:
(1)~the texts of all articles about $E$ in all languages ($\tgtLang$ as well as others),
(2)~the already\hyp existing descriptions of $E$ in all languages other than $\tgtLang$, and
(3)~information about $E$'s semantic type obtained from the Wikidata knowledge graph \cite{vrandecic2014}.

\xhdr{Results}
An automated evaluation on 25 (high- as well as low-resource) languages shows that \name outperforms baselines by a wide margin, including a strong baseline built on state-of-the-art machine translation.
A single multilingual model rivals the performance of a collection of monolingual models custom-trained for individual languages (implying crosslingual transfer capabilities),
and incorporating already\hyp existing descriptions from other languages and knowledge graph information further increases performance.
Given the inherent limitations of automated evaluation metrics for language generation tasks such as ours, we also conducted a crowdsourcing\hyp based evaluation with human raters, who compared \name's descriptions to human\hyp generated gold descriptions of the respective articles.
Across three languages (English, Hindi, Romanian), the quality of our \desc{}s is indistinguishable from, or comes close to, that of the gold \desc{}s.
Manual error analysis of the English case further shows that the errors made by \name follow a similar distribution as those made by human editors, and even when our \desc was not preferred by human raters, it still constitutes a high-quality \desc that could be added to \WP in 60\% of cases (a similar rate as \textit{vice versa,} \ie, for human \desc{}s not preferred by human raters).
Overall, 91.3\% of \name's English \desc{}s (\vs\ 92.1\% of gold \desc{}s) meet \WP's quality criteria, which suggests that our model is ready for production.


\xhdr{Contributions}
In a nutshell, our contributions are the following.

\begin{enumerate}
\item We introduce the novel task of short\hyp \desc generation for \WP articles.
\item We propose \name, a model that tackles the task by merging multilingual textual information and language\hyp independent semantic type information (\Secref{sec:method}).
\item In extensive evaluations we demonstrate that the quality of \name's \desc{}s is largely indistinguishable from human\hyp crafted gold \desc{}s (\Secref{sec:eval auto} and \ref{sec:eval human}).
\end{enumerate}


\begin{table}[]
\caption{Statistics of the 25 \WP language editions considered here. Description length counts tokens for all languages except those marked by *, where it counts characters.
}
\resizebox{\columnwidth}{!}{
\begin{tabular}{@{}llrrrr@{}}
\toprule
   & \textbf{Language} & \textbf{Articles} & \textbf{Missing desc.}\ & \textbf{Missing desc.\ (\%)} & \textbf{Avg.\ desc.\ length} \\ \midrule
en & English           & 5204K             & 1023K                 & 19.65                      & 4.25                        \\
de & German            & 2041K             & 389K                  & 19.07                      & 3.65                        \\
nl & Dutch             & 1886K             & 192K                  & 10.18                      & 4.05                        \\
es & Spanish           & 1463K             & 690K                  & 47.21                      & 3.87                        \\
it & Italian           & 1287K             & 465K                  & 36.14                      & 3.91                        \\
ru & Russian           & 1406K             & 960K                  & 68.25                      & 4.84                        \\
fr & French            & 979K              & 298K                  & 30.45                      & 3.75                        \\
zh & Chinese           & 1025K             & 876K                  & 85.46                      & * 7.38                        \\
ar & Arabic            & 986K              & 315K                  & 31.97                      & 4.07                        \\
vi & Vietnamese        & 122K              & 1172K                 & 95.85                      & 5.72                        \\
ja & Japanese          & 1103K             & 858K                  & 77.78                      & * 13.51                       \\
fi & Finnish           & 451K              & 300K                  & 66.66                      & 2.66                        \\
ko & Korean            & 422K              & 376K                  & 89.11                      & * 14.64                       \\
tr & Turkish           & 321K              & 253K                  & 79.04                      & 4.25                        \\
ro & Romanian          & 282K              & 162K                  & 57.48                      & 4.38                        \\
cs & Czech             & 178K              & 85K                   & 47.89                      & 3.72                        \\
et & Estonian          & 195K              & 160K                  & 81.80                      & 2.50                        \\
lt & Lithuanian        & 185K              & 176K                  & 95.11                      & 1.89                        \\
kk & Kazakh            & 220K              & 219K                  & 99.67                      & 4.00                        \\
lv & Latvian           & 92K               & 71K                   & 77.82                      & 3.65                        \\
hi & Hindi             & 130K              & 80K                   & 61.19                      & 6.34                        \\
ne & Nepali            & 29K               & 25K                   & 85.64                      & 4.53                        \\
my & Burmese           & 44K               & 38K                   & 87.45                      & 2.56                        \\
si & Sinhala           & 17K               & 16K                   & 94.05                      & 3.28                        \\
gu & Gujarati          & 29K               & 7K                    & 25.59                      & 4.84                        \\ \bottomrule
\end{tabular}
}
\label{tab:wp-stats}
\end{table}

\section{Related work}
\label{sec:background}
\begin{figure*}[t]
    \centering
    \includegraphics[width=\textwidth]{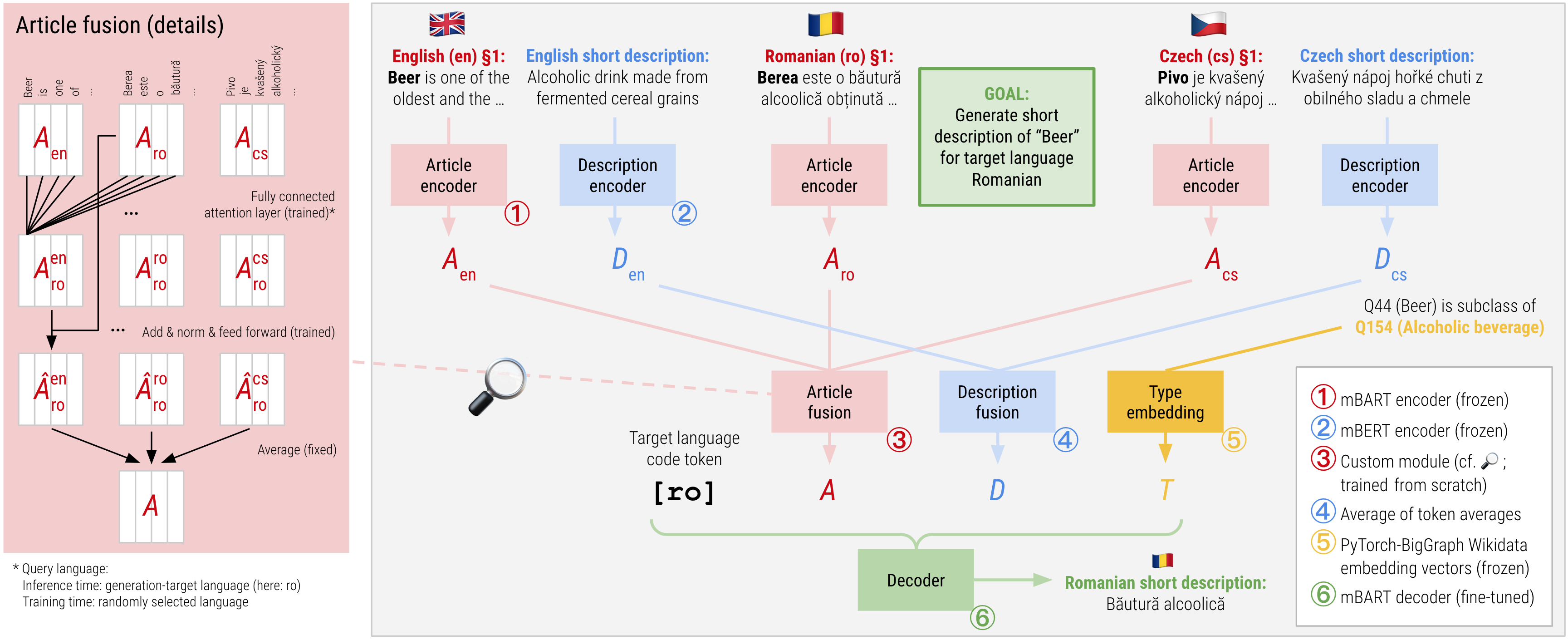}
    \caption{%
Overview of \name, our system for generating \sd{}s for \WP articles, as described in \Secref{sec:method},
illustrated on the task of generating a Romanian \desc of the article about \cpt{Bere} (\cpt{Beer}).
``\S{}1'' stands for ``first paragraph''.
}
    \label{fig:method overview}
\end{figure*}

\xhdr{Wikipedia summarization}
While we are, to the best of our knowledge, the first to tackle the task of short-\desc generation for \WP articles, researchers have previously attempted to automatically generate various other components of \WP articles, including
infoboxes \cite{saez2018, lange2010, yus2014},
section-based article structures \cite{Piccardi_2018},
section titles \cite{field-etal-2020-generative},
keyphrases~\cite{liu-shan2016},
article summaries \cite{lebret-etal-2016-neural,chisholm-etal-2017-learning, vougioklis20181,Vougiouklis2020PointAT},
or entire articles
\cite{sauper-barzilay-2009-automatically, liu2018generating, banerjee-mitra-2015-wikikreator, fan-gardent-2022-generating}.
Unlike ours, most prior work has focused on English \WP, with notable exceptions relevant to our work including multilingual article summarization methods for low-resource languages \cite{kaffee-etal-2018-learning, soton449718}.

\xhdr{Extreme text summarization}
Our work is also related to the area of extreme text summarization, where researchers have aimed at generating one-sentence summaries of news \cite{narayan2018dont, zhang2020pegasus, lewis-etal-2020-bart} or scientific \cite{cachola2020tldr} articles,
or function names for code snippets \cite{pmlr-v48-allamanis16}.
Although these research goals are similar to ours, we emphasize that, by simultaneously leveraging multiple input modalities and languages (\cf\ \Secref{sec:introduction}), our method goes beyond conventional summarization methods.
%
Previous extensions to the standard text summarization paradigm include
multi-document summarization \cite{fan-etal-2019-using, sekine-nobata-2003-survey, 10.1145/3529754, McKeown1995GeneratingSO};
methods that enrich input texts with named-entity labels, part-of-speech tags \cite{nallapati2016abstractive}, or background information sourced from knowledge graphs \cite{huang2020knowledge, fan-etal-2019-using, fernandes2021structured};
and methods that enforce factual consistency in summaries \cite{zhu2021enhancing, cao2017faithful, gunel2020mind}.
In designing \name, we were partly inspired by the design principles behind such methods, but go beyond by tailoring our model to the particularities of the task of short\hyp \desc generation for \WP.

\section{Method}
\label{sec:method}

We begin with a high-level sketch of our method (see overview diagram of \Figref{fig:method overview}) and give further details in the following paragraphs.
As mentioned, \name integrates three input modalities in order to generate \sd{}s for an entity $E$ in language $\tgtLang$:
\begin{enumerate}
    \item \textbf{Article texts} about $E$ in all languages: Articles in different languages may contain different cues, some more relevant than others for generating \sd{}s, and pooling information across languages allows for learning to dynamically select the most relevant information. Note that the mapping from entities to articles is available via Wikidata.
    \item \textbf{Existing descriptions} of $E$ in all languages other than $\tgtLang$: Intuitively, translating from other languages should yield good \desc{}s, and examples in several languages provide even more cues. As different \WP language editions have different norms about \sd{}s, the model should learn how to best leverage each language.
    \item \textbf{Semantic type information:} \Sd{}s typically capture semantic types (\eg, \cpt{Beer} has type \cpt{Alcoholic beverage} in Wikidata), so types and \descs are expected to contain rich cues about one another.
\end{enumerate}

To process these cues, \name starts by transforming raw article texts into distributed (matrix-shaped) representations $A_{\lang}$ for each language $\lang$ in which $E$ has an article.
The language\hyp specific representations $A_{\lang}$ are then fused into a language\hyp independent article representation $A$.
Analogously, raw description texts are transformed into distributed representations $D_{\lang}$, which are then fused into a language\hyp independent description representation $D$.
The semantic type of $E$ is represented via an embedding vector $T$ obtained from the Wikidata knowledge graph.
The description (a token sequence) in target language $\tgtLang$ is then generated by a decoder that receives as input the concatenation of $A$, $D$, $T$, and a special target language token.
This way, \name can flexibly generate \desc{}s for any entity $E$ in any target language $\tgtLang$, whether $\tgtLang$ already has an article for $E$ or not.
The model is trained to maximize the likelihood of $E$'s ground-truth \desc in language~$L$.

The next paragraphs provide more details about the above steps.

\xhdr{Integrating article texts across languages}
Distributed language\hyp specific article representations $A_\lang$ are obtained by feeding raw article texts to the pretrained encoder of the multilingual BART (mBART) language model \cite{liu-etal-2020-multilingual-denoising}.
Here we consider only the first paragraph of each article, where all the required information tends to be contained.
Language\hyp specific article representations $A_\lang$ are then fused into a language\hyp independent representation $A$ as follows (\cf\ red box labeled ``Article fusion (details)'' in \Figref{fig:method overview}):
First, we pick a \emph{query language} $q$ (see below regarding choice of $q$) and use a transformer\hyp style attention mechanism \cite{vaswani2017attention} to fuse $A_q$ in a pairwise fashion with each language $\lang$'s article representation $A_\lang$, thus obtaining a pairwise representation $A_q^\lang$ for every language $\lang$.
Following the standard transformer attention paradigm \cite{vaswani2017attention}, we then add the original $A_q$ to every $A_q^\lang$ (a so-called skip connection), followed by layer normalization and a feed-forward layer, obtaining $\hat A_q^\lang$.
Finally, we average $\hat A_q^\lang$ over all languages $\lang$ in order to obtain the language\hyp independent article representation $A$.
(For a more formal description, \cf\ \Appref{appendix:model_desc}.)

During training, we use a randomly sampled language as the query language $q$, which leads to better generalization than when always using the target language $L$ as $q$; \eg, it enables using target languages at inference time that do not even have an article about entity $E$ yet.
This said, we found that, during inference, using the target language $L$ as $q$ gives the best results when $L$ has an article about $E$, so we choose $q=L$ in such cases.

\xhdr{Integrating existing \desc{}s across languages}
Distributed language\hyp specific representations $D_\lang$ of already\hyp existing \desc{}s are obtained by feeding raw description texts to the pretrained encoder of the multilingual BERT (mBERT) language model%
\footnote{
\label{fn:mBERT}
\url{https://huggingface.co/bert-base-multilingual-uncased}
}
and then averaging all token embeddings into a single vector.
In order to obtain the language\hyp independent \desc representation $D$, we then average $D_\lang$ over all languages $\lang$.
We opted for simple averages over tokens and languages because this was found to be as effective as more sophisticated aggregation mechanisms during development.
Similarly, we chose mBERT instead of the above\hyp introduced mBART because it is a smaller model but performed well during development.

\xhdr{Integrating semantic types}
Type information for \WP articles is readily available from the Wikidata knowledge graph \cite{vrandecic2014}, which links each language\hyp specific article (\eg, \cpt{Beer}) to a language\hyp independent entity (\eg, Q44) and specifies semantic types (\eg, Q154, \ie, \cpt{Alcoholic beverage}) for almost all entities via the \cpt{instance of} and \cpt{subclass of} relations.
To obtain distributed type representations that can be plugged into our neural network model, we use PyTorch-BigGraph's%
\footnote{
\label{fn:PBG}
\url{https://github.com/facebookresearch/PyTorch-BigGraph}}
\cite{pbg}
Wikidata graph embedding (which maps every Wikidata entity to a 200\hyp dimensional vector) and use as $E$'s type representation $T$ the average embedding vector of all Wikidata entities $E'$ for which the relation $(E, \text{\cpt{instance of}}, E')$ or $(E, \text{\cpt{subclass of}}, E')$ holds (or, if there is no such $E'$, a dummy representation equal to the average over all type embedding vectors).

\xhdr{Implementation}
We implemented the above model (available as per footnote~\ref{fn:code and data}) in PyTorch on top of Hugging Face.
The mBART model%
\footnote{\url{https://huggingface.co/facebook/mbart-large-cc25}}
supports 25 languages and has 610M parameters (12-layer encoder, 12-layer decoder, 16 heads, 1024\hyp dimensional hidden layers).
The mBERT model (\cf\ footnote~\ref{fn:mBERT}) supports 102 languages and has 110M parameters (12-layer encoder, 12 heads, 768\hyp dimensional hidden layers).
As mBART's 25 languages form a subset of mBERT's 102 languages, our implementation of \name supports those 25 languages (\cf\ \Tabref{tab:wp-stats}),
which span many language families and the full spectrum from low to high resources.

The pretrained encoders of mBART and mBERT, as well as the PyTorch-BigGraph entity embeddings, are frozen and used without fine-tuning,
whereas the mBART decoder is finetuned during training
and the article fusion module (red box in \Figref{fig:method overview}) is trained from scratch.
This way, training remains efficient, with only 215M trainable parameters, compared to 720M for full mBART and mBERT combined.
Finetuning on 100K \descs took 24 hours on a single GPU, and generating all 9.3M \descs currently missing in the 25 supported languages would take around 40 days on a single GPU (but note that the process can be trivially parallelized and needs to be run only once).

\section{Automatic evaluation}
\label{sec:eval auto}

\begin{figure}
  \centering
  \includegraphics[width=\columnwidth]{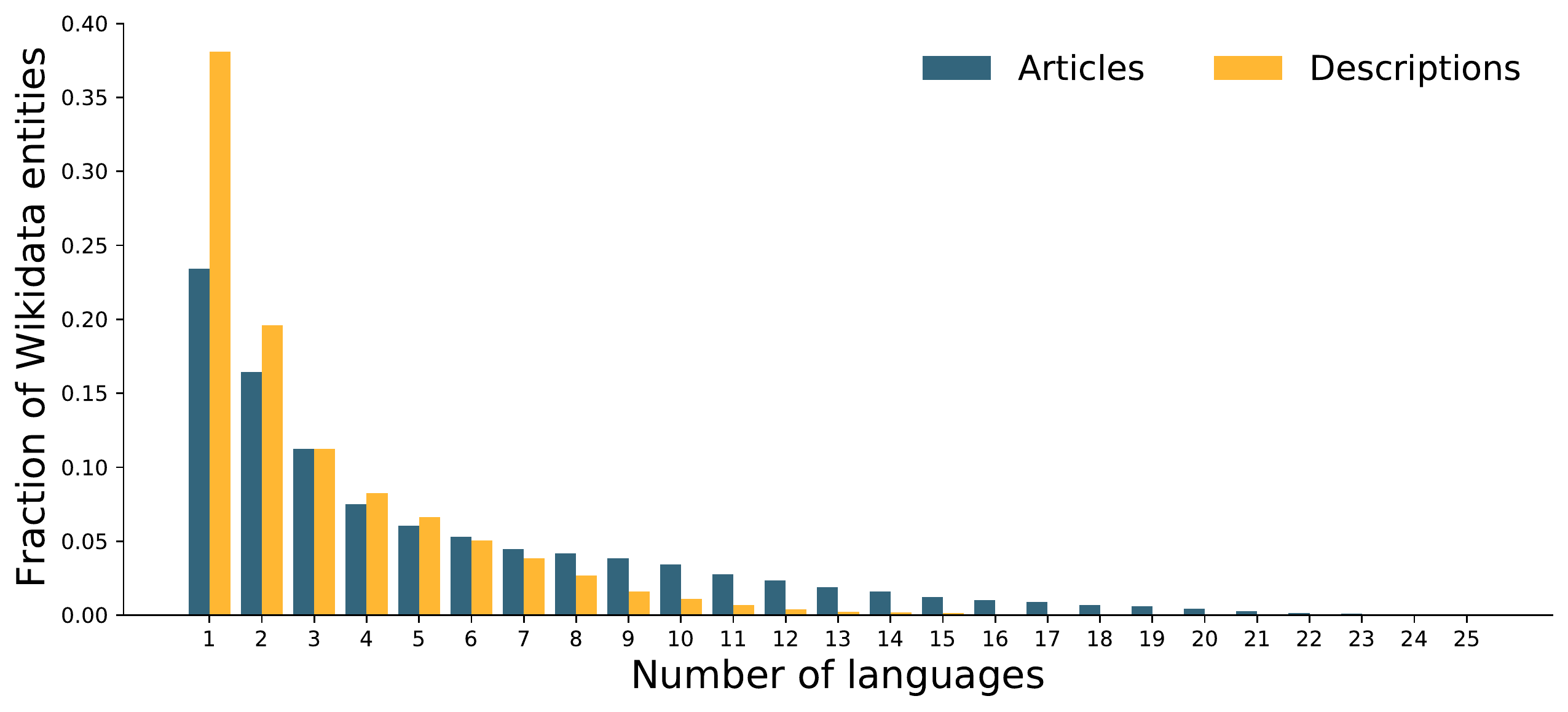}
  \vspace{-7mm}
  \caption{%
Distribution of number of languages in which Wikidata entities have articles (blue) or descriptions (orange).}
  \label{fig:distribution}
\end{figure}

\begin{table*}[t]
\centering
\caption{
Automatic evaluation in 25 languages (\cf\ \Tabref{tab:wp-stats})
via average MoverScores.
First row shows percentage of Wikidata entities with article in language.
Notes (\cf\ \Secref{sec:auto eval setup}):
Translation baseline evaluated only on a subset of the test set (articles with a description in at least one other language; 61.9\% of test instances).
Summarization baseline (evaluated only for English): .537.
}
\resizebox{\textwidth}{!}{
\setlength{\tabcolsep}{3pt}
\begin{tabular}{@{}lccccccccccccccccccccccccc@{}}
\toprule
& en & de & nl & es & it & ru & fr & zh & ar & vi & ja & fi & ko & tr & ro & cs & et & lt & kk & lv & hi & ne & my & si & gu \\
\midrule
\textbf{\% entities w/ art.}   & 77 & 45 & 41 & 41 & 37 & 36 & 35 & 27 & 27 & 24 & 24 & 17 & 16 & 14 & 12 & 9 & 7 & 7 & 7 & 5 & 4 & 1 & 1 & .7 & .7 \\
\midrule
\multicolumn{3}{l}{{\textbf{Baselines}}} \\ 
\hspace{4mm} Prefix  & .564 & .563 & .572 & .565 & .558 & .622 & .556 & .627 & .651 & .581 & .594 & .584 & .618 & .581 & .560 & .567 & .596 & .616 & .622 & .570 & .636 & .645 & .710 & .750 & .785 \\
\hspace{4mm} Translation  & .682 & .661 & .619 & .686 & .647 & .704 & .688 & .693 & .716 & .704 & .679 & .661 & .703 & .658 & .648 & .659 & .775 & .693 & .665 & .674 & .733 & .688 & .713 & .903 & .872  \\

\midrule
\multicolumn{3}{l}{{\textbf{Proposed models}}} \\
\hspace{4mm} \name  & \textbf{.781} & \textbf{.798} & \textbf{.846} & .838 & .855 & .821 & \textbf{.825} & .815 & .883 & .818 & .840 & .814 & .802 & .704 & .900 & \textbf{.807} & .888 & .882 & .633 & .747 & .888 & \textbf{.800} & .835 & .862 & .900 \\
\hspace{4mm} [no desc]  & .778 & .793 & \textbf{.846} & .834 & .853 & .824 & .816 & .812 & .879 & .825 & .841 & .814 & \textbf{.810} & .696 & .898 & .787 & .886 & .874 & .604 & .773 & \textbf{.890} & .788 & .795 & .821 & .895  \\
\hspace{4mm} [no types]  & .775 & .794 & .844 & \textbf{.840} & .852 & .819 & .814 & .815 & \textbf{.884} & .831 & .842 & \textbf{.816} & .803 & \textbf{.709} & .898 & .794 & \textbf{.898} & .872 & \textbf{.653} & .756 & .891 & .780 & \textbf{.836} & .901 & .899 \\
\hspace{4mm} [no desc/types]  & .772 & .776 & .840 & .836 & .847 & .824 & .810 & .809 & .881 & \textbf{.837} & .833 & .812 & .800 & .697 & .894 & .794 & .881 & \textbf{.892} & .614 & .753 & .888 & .771 & .772 & .923 & .892 \\
\hspace{4mm} [monolingual]  & .770 & \textbf{.798} & .833 & .835 & \textbf{.857} & \textbf{.827} & .810 & \textbf{.820} & .881 & .827 & \textbf{.850} & \textbf{.816} & .794 & .708 & \textbf{.907} & .793 & .887 & .889 & .591 & \textbf{.800} & .889 & .791 & .810 & \textbf{.935} & \textbf{.978} \\

\bottomrule                            
\end{tabular}
}

\label{tab:eval-systems}
\end{table*}

To quantify \name's performance, we start with a large-scale evaluation based on automatically computed scores measuring the similarity between auto\hyp generated and human\hyp generated \desc{}s.
Here the latter serve as a gold standard, and perfect performance corresponds to reproducing the human\hyp generated ground truth exactly.
This evaluation is imperfect since
(1)~automatically computed scores may not reflect actual output quality perfectly and
(2)~it cannot correctly score cases where auto\hyp generated \desc{}s are better than the human\hyp generated ones, which are taken as perfect by defintion.
We therefore later complement our analysis with a human evaluation where \name's \desc{}s are explicitly compared against human\hyp generated ones (\Secref{sec:eval human}).

\subsection{Experimental setup}
\label{sec:auto eval setup}

\xhdr{Data}
The automatic evaluation was conducted for the 25 languages of \Tabref{tab:wp-stats}, based on three sets of Wikidata entities sampled uniformly at random under the constraint that all sampled entities have an article and a \sd in at least one of the 25 languages:
a training set of 100K entities,
a testing set of 10K entities,
and a validation set (used during development) of 10K entities.
During training, each entity from the training set contributed one data point, with a randomly sampled language (out of those for which a description was available) serving as the target language.
During testing, we generated \desc{}s for all languages in which the respective entity had a human\hyp generated ground-truth \desc.

\Figref{fig:distribution} shows the distribution of the number of languages in which Wikidata entities from the training set have articles and \desc{}s.
Although entities with an article in only one language are most common, 77\% have articles in at least two languages.
Similarly, although entities with a \desc in only one language are most common, 62\% have \desc{}s in at least two languages (one of which serves as a target language during training).
Additionally, for 94\% of entities, a semantic type could be extracted from Wikidata (\cf\ \Secref{sec:method}).
These statistics show that all three modalities considered by \name (multilingual article texts, multilingual descriptions, semantic types) can be exploited in the majority of cases.

For a sample of descriptions by \name, see~\Appref{appendix:results}.

\xhdr{Models}
In our analysis we compare eight methods:
full \name as described in \Secref{sec:method},
four ablated versions that ignore one or more input modalities,
and three baselines:
\begin{enumerate}
\item \textbf{\name:}
Our full model (\cf\ \Secref{sec:method}).
\item \textbf{\name [no desc]:}
Ignoring existing descriptions.
\item \textbf{\name [no types]:}
Ignoring semantic type information.
\item \textbf{\name [no desc/types]:}
Using only (multilingual) article texts.
\item \textbf{\name [monolingual]:}
Using only the article text in the target language. Note that this requires training 25 models, one per target language.
\item \textbf{Prefix baseline:}
Sanity-check baseline that returns the first $n$ words of the article (in the target language) for which a \desc is to be generated, where $n$ is the average number of words in descriptions in the target language (\cf\ rightmost column of \Tabref{tab:wp-stats}; for Chinese, Japanese, and Korean, characters were used as a unit instead of words).
\item \textbf{Translation baseline:}
When descriptions are available in other languages, we translate one to the target language using Google Translate. If more than one are available, we choose the highest\hyp resource language (in terms of number of \WP articles). If no descriptions in other languages are available, this baseline cannot be applied.
\item \textbf{Summarization baseline:} BART \cite{lewis-etal-2020-bart} finetuned on the XSum extreme summarization dataset \cite{narayan2018dont} (English only).
\end{enumerate}

\xhdr{Performance metric}
As the main performance metric for automatic evaluation
we use MoverScore~\cite{zhao-etal-2019-moverscore},
which was designed for measuring the semantic similarity between auto\hyp generated text and a ground-truth reference.
The MoverScore lies in $[0,1]$, and larger values are better.
It leverages multilingual contextual representations and is based on the earth mover's distance in embedding space between the two texts.
By working in semantic\hyp embedding space rather than surface\hyp token space, MoverScore correlates better with human judgment than token\hyp matching metrics such as BLEU or ROUGE, as was shown on a variety of text generation tasks such as summarization or translation \cite{zhao-etal-2019-moverscore}.
This fact is particularly important in abstractive settings such as ours, where many good outputs consisting of entirely different words are possible.

\subsection{Results}
\label{sec:auto eval results}

\Tabref{tab:eval-systems} shows the test-set performance (average MoverScore) of all eight evaluated methods.
We see that \name beats the baselines in each language, whereas
among the \name versions, there is no clear winner across languages.
Full \name performs best in 6 languages,
\name [no types] in 7,
\name [no desc] in 3,
\name [no desc/types] in 2,
and the respective monolingual version optimized for the respective target language in 10 languages.
(The sum is greater than 25 due to ties.)
The fact that the multilingual \name versions perform similarly to, or even better than, the monolingual versions across languages is particularly encouraging, as it implies that there is little to no performance decline in the multilingual setting and that, consequently, we can cater to both high- and low\hyp resource languages with one unified model.

Given that translating an existing \desc from another language \textit{a priori} appears to be a powerful heuristic, it may be surprising that \name outperforms the translation baseline by such large margins (\eg, by 15\%\slash 21\%\slash 37\% in English\slash German\slash Dutch, the largest languages in our data).
However, different languages often have different conventions regarding \desc styles, and machine\hyp translating a text consisting of only a few words might be an under\hyp constrained problem.
Both issues are alleviated by \name: as a supervised model, it can learn language\hyp specific conventions,
and article texts as well as multiple existing \desc{}s can provide disambiguating contexts that further constrain the problem.

Whereas the above column\hyp wise comparisons in \Tabref{tab:eval-systems} (of methods for a given language) are valid, row-wise comparisons and averages (of languages for a given method) are more problematic, as MoverScores may not be comparable across languages for various reasons.
For instance, scores in low\hyp resource languages may be inflated because embeddings of texts in these languages tend to be closer to one another in the mulitlingual embedding space, which may bias their mutual distances to smaller values.
Additionally, Wikipedia articles in low-resource languages cover a narrower range of topics, which makes the \desc{}\hyp generation task inherently easier for these languages and may inflate MoverScores.

\xhdr{Pairwise comparison}
To circumvent these issues, we deploy a more robust, paired aggregation methodology termed \emph{Pairformance} \cite{peyrard-etal-2021-better},
which uses the Bradley--Terry model \cite{BT} to
infer a latent performance score $s(m)$ for each model $m$ such that the probability that model $m_1$ performs better than model $m_2$ on a randomly drawn test instance (across all languages) is approximated by
$\frac{s(m_1)}{s(m_1)+s(m_2)}$.
This aggregation is scale\hyp independent,
which is important in our case due to the non\hyp comparability of absolute MoverScores across languages.
\Tabref{tab:bt-comp-full}, which contains a comparison of all models as captured by the above pairwise probabilities,%
\footnote{The translation baseline is omitted from the table as it cannot handle test instances without \desc{}s in other languages (\cf\ \Secref{sec:auto eval setup}).}
demonstrates that leveraging existing descriptions and semantic types increases model performance statistically significantly, compared to a model that uses only article texts (\name [no desc\slash types]);
\eg, full \name beats \name [no desc\slash types] on 52.2\% of test instances,
while \name [no desc] and \name [no types] do so on 51.5\% and 52.1\%, respectively.
These results also confirm that, across languages, the unified multilingual \name model performs nearly indistinguishably from a collection of language\hyp specific monolingual models (the latter having a win ratio of 50.5\%).

The fact that different versions of \name perform similarly is partly due to the fact that, as we show in \Secref{sec:eval human}, the model has little room for improvement, given that it is already indistinguishable from human performance.
Moreover, the fact that adding existing descriptions to the model gives a similar boost as adding semantic types may be caused by these two cues carrying similar information.
By manually inspecting model outputs, we observed, however, that each of these signals can improve the generated \desc{}s in certain individual cases (examples in \Appref{appendix:results}).


\begin{table}[]
\caption{
\boldmath
Pairwise automatic model comparison.
Cell $(i,j)$ has probability that model of row $i$ beats model of column $j$ on random test instance, as estimated via Bradley--Terry model (\cf\ \Secref{sec:auto eval results}).
Asterisks (*) mark probabilities statistically significantly ($p<0.05$) different from 0.5 under sign test.
\unboldmath}
\resizebox{\columnwidth}{!}{
\begin{tabular}{l|c|c|c|c|c|c}
                     & Prefix & \name & {[}no desc/types{]} & {[}no desc{]} & {[}no types{]} & {[}monolingual{]} \\ \hline
Prefix     &                 & 0.075*               & 0.077*              & 0.076*        & 0.076*        & 0.076*            \\ \hline
\name & 0.925*          &                      & 0.522*              & 0.508*        & 0.501*        & 0.495*            \\ \hline
{[}no desc/types{]}  & 0.923*          & 0.478*               &                     & 0.485*        & 0.479*        & 0.490              \\ \hline
{[}no desc{]}        & 0.925*          & 0.492*               & 0.515*              &               & 0.493*        & 0.492*            \\ \hline
{[}no types{]}        & 0.924*          & 0.499*               & 0.521*              & 0.507*        &               & 0.496*            \\ \hline
{[}monolingual{]}    & 0.924*          & 0.505*               & 0.510                & 0.508*        & 0.504*        &                   \\
\end{tabular}
}
\label{tab:bt-comp-full}
\end{table}

\begin{figure}
  \centering
  \includegraphics[width=0.9\columnwidth]{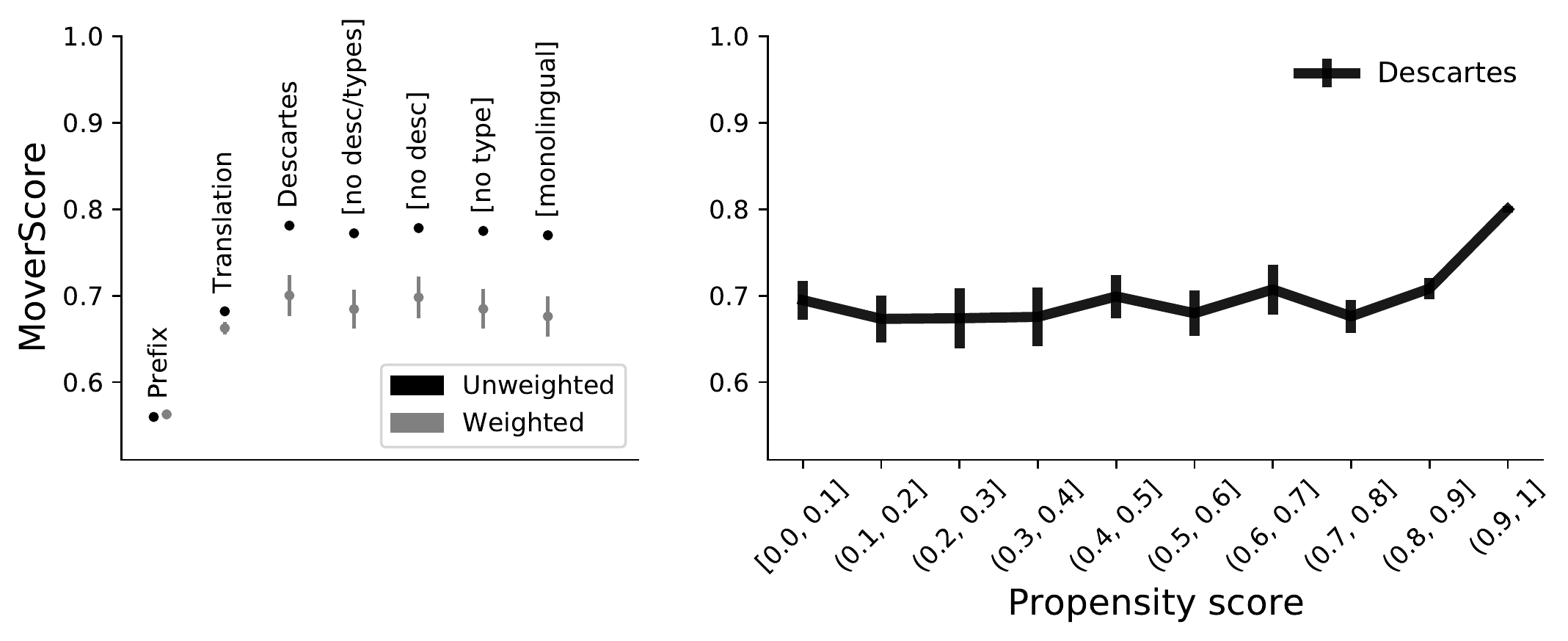}
  \vspace{-3mm}
  \caption{
Propensity-score-based analysis analysis of automatic evaluation on English test set.
\emph{Left:}~Unweighted (black, \cf\ \Tabref{tab:eval-systems}) and propensity\hyp score\hyp weighted (gray, for approximating distribution of articles without \descs) average MoverScores.
\emph{Right:}~MoverScore of \name stratified by propensity score.
Error bars: 95\% confidence intervals (black error bars in left plot are too small to be visible).
}
  \label{fig:propensity}
 \vspace{-3mm}
\end{figure}

\xhdr{Propensity-score-based analysis}
\name's use case is to generate descriptions for articles that do not have one yet.
For evaluation, we, however, needed to use articles that already have a description, which may differ systematically from articles that do not have a description yet (\eg, they may be about more popular topics, better written, \etc).
If this is the case, our test set would not be fully representative of the articles to which our model will be deployed, and the above\hyp reported performance would be a biased estimate.

To correct for this potential mismatch, we employ \textit{propensity scores} \cite{austin2011introduction}, a tool commonly used in observational studies in order to correct for such biases.
In our case, an article $i$'s propensity score $p_i$ is defined as its probability of already having a \sd, judging by its textual content only.
Propensity scores allow us to reweight \cite{Lunceford2004StratificationAW} the test set, which by construction contains only articles that already have a \desc, in order to reflect the distribution of articles that do not have a \desc yet.
In particular, by defining article $i$'s weight as $(1-p_i)/p_i$ and taking a weighted average of MoverScores over the test articles allows us to estimate the expected performance for articles without \desc{}s, even though we cannot directly evaluate on such articles.

Focusing on English, we created a dataset containing randomly sampled English articles, of which around 20\% have no \desc (\cf\ \Tabref{tab:wp-stats}), and finetuned a BERT-based \cite{devlin-etal-2019-bert} binary classifier for predicting if an article $i$ has a \desc, based on $i$'s text.
The classifier's softmax output then serves as $i$'s propensity score $p_i$.

\Figref{fig:propensity} (left) summarizes the performance of all methods via weighted as well as unweighted average MoverScores (the latter stemming from \Tabref{tab:eval-systems}).
Although all scores are lower in the weighted case, implying that test articles that resemble articles without \desc{}s (which are upweighted in this analysis) tend to have lower MoverScores,
the ordering of methods remains the same as in the unweighted analysis.
\Figref{fig:propensity} (right), which plots the average MoverScore of full \name stratified by propensity score, further shows that test performance is stable across the full propensity\hyp score range, with only those test articles with the highest propensity scores standing out with higher-than-average MoverScores.
This implies that \name works well across the full range of articles.

\xhdr{Limits of automatic evaluation}
In interpreting the results of the above automatic evaluation, an important caveat must be kept in mind:
low\hyp propensity\hyp score articles can be expected to also have lower\hyp quality human\hyp generated ground-truth \descs,
but the evaluation metric (MoverScore) measures the similarity between auto\hyp generated and ground-truth \descs.
Hence, low MoverScores may be due to low-quality model outputs or due to low-quality ground truth.
Moreover, even if the ground truth were perfect, the automatic metric might not reflect similarity to it perfectly,
and automatic metrics are only useful for comparing relative improvements between models, rather than measuring whether the outputs are ``good enough'' to solve the task.
In order to overcome these fundamental limitations of any automatic evaluation, we conducted an evaluation with human raters, described next.


\section{Human evaluation}
\label{sec:eval human}
To surpass the above limitations of automatic metrics, we conducted a human evaluation on Amazon Mechanical Turk (MTurk).
Whereas in the automatic evaluation the human\hyp generated \desc was a gold standard, here it is a competitor, so the auto\hyp generated \desc is now able to score higher than the human\hyp generated one, which was impossible in the automatic evaluation.
We evaluated \name (in its full version) on three languages: English, Hindi, and Romanian.
These languages were chosen because they span different language families and have different amounts of text available for training,
while all being spoken by a large number of MTurk workers~\cite{pavlick-etal-2014-language}, which facilitated recruiting.


\subsection{Experimental setup}
\label{sec:human eval setup}

\xhdr{Task design}
In each rating task, workers were shown an entire Wikipedia page alongside the auto- and the human\hyp generated \descs (in random order to avoid positioning bias), and were asked to choose the more appropriate one, in a forced-choice manner, following standard methodological recommendations \cite{PrezOrtiz2017APG}.
The instructions stated that ``a good \desc indicates in one phrase what the article is about''
and listed examples of good descriptions borrowed from \WP's guidelines (\cf\ footnote~\ref{fn:SD instructions}).
Each MTurk task consisted of a batch of 10 binary choices (each for a different \WP article).
One of these was a fabricated honeypot, where the auto\hyp generated \desc was replaced with the ground-truth \desc of a randomly sampled different article, such that it was obvious which of the two \descs was better.
This allowed us to filter out unreliable workers \textit{post hoc} (see below).
Each rating task was done by three different workers, and the winning \desc per pair was determined by majority vote.

\xhdr{Data}
To avoid comparisons between two identical descriptions, we removed from the test set all the articles for which \name reproduced the ground truth (up to capitalization),
which eliminated 37.5\%, 64.6\%, 69.0\% of articles in English, Hindi, and Romanian, respectively
(itself a manifestation of the quality of \name's output).
From the rest, we sampled a total of 1000 articles per language, stratifying by MoverScore and sampling 100 articles per decile, with the intent to cover the full performance range according to the automatic evaluation and to thus be able to determine whether the automatic evaluation scores correlate with human choice.
For English, sampling within each MoverScore decile was uniformly at random.
As Romanian and Hindi \WP have less topical diversity, we drew biased samples with the goal of covering the topical spectrum more evenly, by representing articles via their Wikidata graph embedding (\cf\ footnote~\ref{fn:PBG}) and sampling far-apart data points using the k-means++ \cite{ilprints778} cluster seeding algorithm.

Since, to enforce conciseness, we limited \name's output length when generating \descs,
some (in particular for disambiguation pages) ended abruptly, before the decoder could terminate its output.
Such test instances were omitted from the analysis.


\xhdr{Crowdworkers}
To ensure reliable ratings, we recruited only workers with at least a minimum number\slash fraction of previously approved tasks (1000/99\% for English; 500/97\% for Hindi and Romanian).
Hindi and Romanian participants were restricted by location (India and Romania, respectively).
We targeted a pay rate of \$8--10 per hour, guided by US minimum wage.
In order to verify that workers understood Hindi or Romanian, respectively, they needed to correctly answer a simple multiple\hyp choice entry question in that language.
Finally, we filtered unreliable workers by excluding those who failed on over 20\% of the encountered honeypots (see above).

\subsection{Results}
\label{sec:human eval results}

\begin{figure}
  \centering
  \includegraphics[width=\columnwidth]{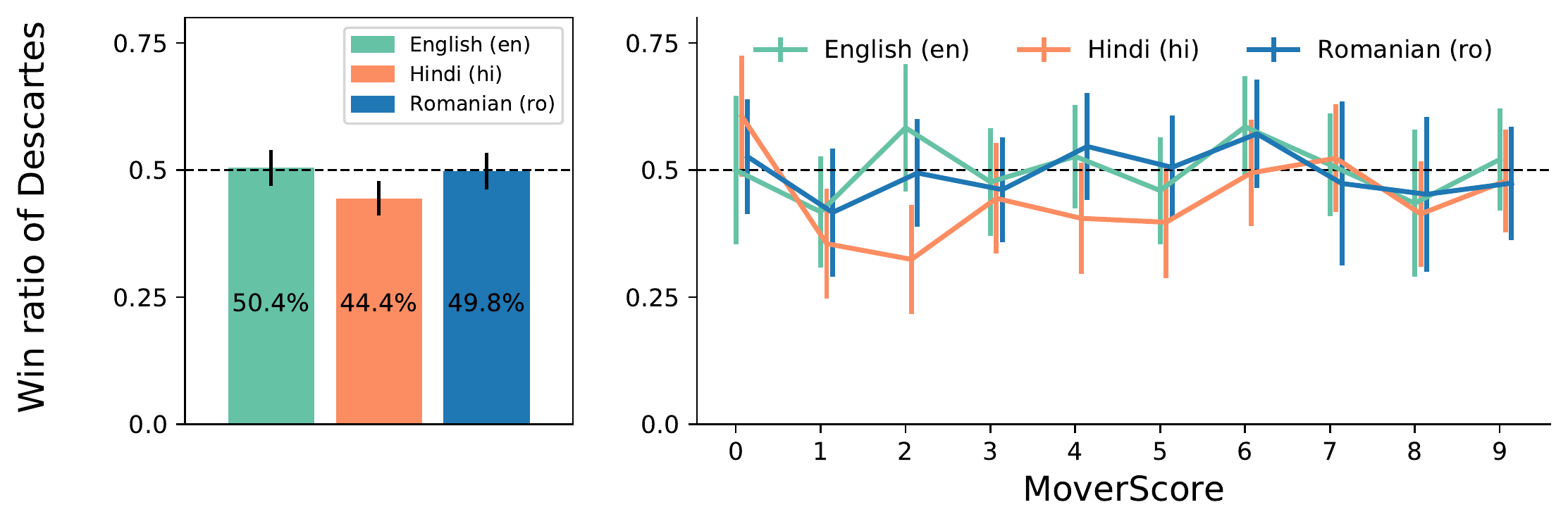}
  \vspace{-7mm}
  \caption{
Human evaluation results, showing that \name's \descs are largely indistinguishable from human\hyp written ground truth.
\emph{Left:}~Fraction of articles for which human raters preferred \name over ground truth.
\emph{Right:}~\textit{Idem,} stratified by automatically computed similarity (MoverScore) of \name and ground truth.
Error bars: 95\% CIs.}
  \label{fig:human_eval}
 \vspace{-3mm}
\end{figure}

\xhdr{Performance analysis}
The results, reported in \Figref{fig:human_eval} (left), show that our \descs were rated as better than their human\hyp generated counterparts in 50.5\% [47.3\%, 53.3\%], 44.4\% [41.6\%, 47.3\%], and 49.8\% [46.7\%, 52.9\%] of the tested articles in English, Hindi, and Romanian, respectively (95\% CIs in brackets; Fleiss' $\kappa =$ 0.23, 0.32, 0.78).
First, this shows that \name performs similarly well for all three languages, despite them having widely different amounts of training data.
Second, as implied by the 95\% CIs containing 50\%, \name's \descs in English and Romanian are of indistinguishable quality from the human\hyp generated ones, whereas for Hindi the latter were only slightly preferred by raters.
\Figref{fig:human_eval} (right), which disaggregates the results by MoverScore decile, shows that performance is high throughout.
The lack of correlation between MoverScores and human preference \textit{post hoc} supports the aforementioned doubts regarding the appropriateness of automatic scores such as MoverScore:
while the similarity with the ground truth is lower in lower MoverScore bins by definition, the ground truth itself seems to be of lower quality there as well.
As a concrete example, in the lowest MoverScore decile, \cpt{Ain O Salish Kendro} (a Bangladeshi NGO) has ground truth \descText{Organization}, whereas \name generated the better \descText{Human rights organization in Bangladesh}.

\begin{table}[tb!]
\caption{Error categories for human evaluation.}
\centering
\resizebox{0.9\columnwidth}{!}{
\setlength{\tabcolsep}{3pt}
\begin{tabular}{l|p{70mm}}
\hline
\textbf{Error category}   & \textbf{Description}     \\ \hline \hline
Too vague        & The description does not contain enough information to identify the entity                                                                                         \\ \hline
Factual error    & The description contains wrong facts, for example a wrong year, nationality of a person                                                                            \\ \hline
Formatting error & The description is not an appropriate short-description independent of its content: it contains grammatical errors, formatting problems like using ``\dots'', errors with white spaces \\ \hline
Mis-focused      & The description is describing other entities                                                                                                                       \\ \hline
Too long         & The description contains too many details, is too long, or made of several sentences                                                                               \\ \hline
\end{tabular}
}
\label{tab:err-cat}
 \vspace{-3mm}
\end{table}

\xhdr{Error analysis}
To further investigate the quality of auto-generated descriptions and their human\hyp generated counterparts, we manually inspected the descriptions (both auto- and human\hyp generated) that were not selected by MTurk raters and categorized them based on an error taxonomy developed via iterative coding (see \Appref{appendix:error taxonomy} for details), shown in~\Tabref{tab:err-cat}.
Note that, due to the forced-choice setting, even a non\hyp preferred \desc may be---and indeed often is---of high quality, in which case it was labeled as ``good enough''.

\begin{figure}
  \centering
  \includegraphics[width=0.9\columnwidth]{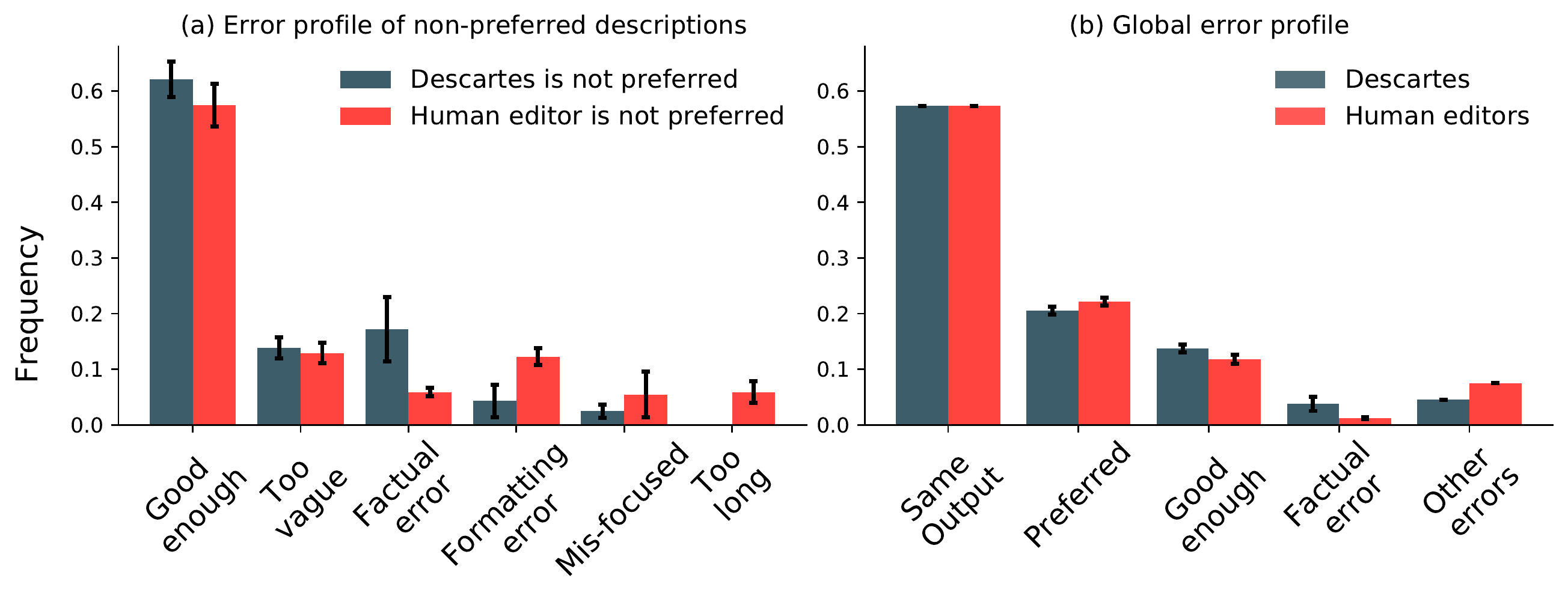}
 \vspace{-5mm}
  \caption{
Error analysis on English test set, for
(\textit{left})~descriptions not preferred by human raters and
(\textit{right})~all descriptions, \ie, including preferred \descs and \descs that were produced identically by \name and human \WP editors.
91.3\% of \name's descriptions are either identical to the ground truth, preferred over the ground truth, or non\hyp preferred but still of high quality (``good enough''), similar to the analogous number for ground-truth \descs (92.1\%).
Error bars: 95\% CIs.
}
  \label{fig:error-analysis}
\end{figure}

Using this taxonomy, two authors annotated the non\hyp preferred description of 300 random test articles, without knowing how it was generated (Fleiss' $\kappa=0.77$; disagreements were manually resolved).
The error distributions for auto- and human\hyp generated \descs, plotted in \Figref{fig:error-analysis} (left), are nearly identical, the main differences being that \name makes more factual errors, but never generates overly long descriptions (as output length was limited during decoding).
Manual inspection further revealed that factual errors resulted from
(1)~generating incorrect years
or
(2)~not using the word ``former'' for outdated properties.
(The second type of error was also commonly found in human\hyp generated descriptions.)

We thus obtain the global error profile of \Figref{fig:error-analysis} (right),
which represents the non\hyp preferred \descs of \Figref{fig:error-analysis} (left) as well as the preferred \descs and cases where auto- and human\hyp generated \descs are identical.
(The rarest error types are grouped as ``other errors'' here.)
Summing up the three leftmost bars, we conclude that 91.3\% of \name's descriptions are either identical to the ground truth, preferred over the ground truth, or non\hyp preferred but still of high quality (``good enough'').
For the human ground truth, the analogous fraction is 92.1\%.
Taken together, this analysis demonstrates that the quality of \name's descriptions is on par with that of human-written \descs.

\begin{figure}
  \centering
  \includegraphics[width=\columnwidth]{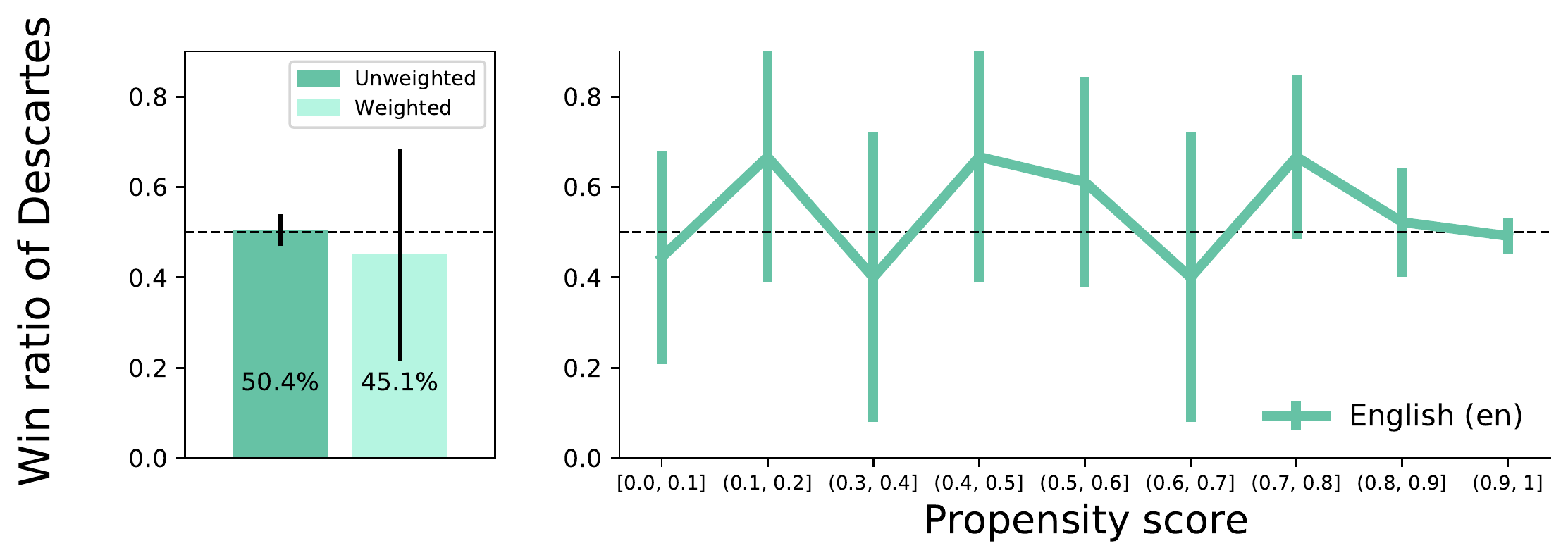}
 \vspace{-6mm}
  \caption{
Propensity-score-based analysis of human evaluation on English test set.
\emph{Left:} Unweighted (dark) and propensity\hyp score\hyp weighted (light) average win ratio of \name (\ie, fraction of articles where human raters preferred \name over ground truth).
\emph{Right:} Win ratio stratified by propensity score.
Error bars: 95\% CIs.
}
  \label{fig:human_eval_prop}
\vspace{-3mm}
\end{figure}

\xhdr{Propensity-score-based analysis}
Finally, we analyze the results using propensity scores (\cf\ \Secref{sec:eval auto}), to investigate possible biases due to the evaluation being done on articles that already have a \desc in the target language, whereas the model is intended for articles that do not have one yet.
As before (\cf\ \Figref{fig:propensity}), we stratify the results by propensity score and compute propensity\hyp score\hyp weighted averages (\Figref{fig:human_eval_prop}), finding that the fraction of times \name's \desc was preferred does not vary systematically with the propensity to have a \desc and is always close to 50\%---further evidence that the model works well on its intended use case: to write \descs for articles that do not have one yet.

\section{Conclusion}
\label{sec:discussion}
This work addresses a large and consequential gap in \WP: that millions of articles across languages have no \sd yet, which hampers the searchability, navigability, and maintainability of the world's largest encyclopedia for millions of users.

Our proposed model---\name---is, to the best of our knowledge, the first solution to this problem in the literature.
Our automatic evaluation demonstrates that a single multilingual model
performs better than a strong machine translation baseline and as good as monolingual models that were specifically optimized for individual languages.
Our human evaluation went further by showing that \name is essentially indistinguishable from human-written \descs across three languages (English, Hindi, Romanian).

Encouraged by these results, we are currently working on turning \name into a practical tool with the goal of supporting human editors in decreasing the otherwise ever\hyp growing number of missing \descs.
We envision a micro-task that can be accomplished even by less experienced editors (perfectly suited as an onboarding exercise for newcomers), \eg, via an interface showing an article snippet alongside \name's top \descs, from which the editor chooses the best (with the option to reject \name's recommendations and instead contribute another one).

More broadly, this work highlights the potential of solutions built upon state-of-the-art generative language models to help close content gaps in \WP that would otherwise keep widening forever, given the scarcity of volunteer editor time.
We are looking forward to seeing more applications in the same spirit.

{\small
\xhdr{Acknowledgments}
With support from
Swiss National Science Foundation (200021\_185043),
H2020 (952215),
Microsoft,
Google,
and Facebook.
}

\bibliographystyle{ACM-Reference-Format}
\bibliography{bibliography}

\appendix

\section{Wikidata descriptions}
\label{appendix:wikidata}

\WP \sd provide a concise indication of the field covered by the article. There exists a similar concept in Wikidata called \emph{Wikidata descriptions,}\footnote{\url{https://www.wikidata.org/wiki/Help:Description}}
whose purpose is to disambiguate Wikidata items with similar labels.
One might be tempted to use Wikidata descriptions as \WP \descs and \textit{vice versa,} but as \Tabref{tab:wd-stats} shows, this cannot solve the problem, as a significant fraction of Wikidata items have no description either---and when they do have a \desc, the corresponding \WP article already does so, too, in the vast majority of cases. This high overlap is indicated by the Jaccard coefficient in the rightmost column of \Tabref{tab:wd-stats}. Additionally, when both Wikipedia and Wikidata \descs are present, they are almost always the same, with the exception of English, where only 58.37\% are identical.

\begin{table}[]
\caption{Statistics of Wikidata items connected to 25 \WP language editions.}
\resizebox{\columnwidth}{!}{
\begin{tabular}{@{}llrrrrr@{}}
\toprule
   & Language   & Articles & Missing desc. & Missing desc.\ (\%) & Exact copies (\%) & Jaccard \\ \midrule
en & English    & 5204K    & 556K         & 10.68             & 58.37           & 0.8904  \\
de & German     & 2041K    & 321K         & 15.71             & 96.75           & 0.9585  \\
nl & Dutch      & 1886K    & 181K         & 9.60              & 98.60           & 0.9913  \\
es & Spanish    & 1463K    & 631K         & 43.17             & 95.16           & 0.9263  \\
it & Italian    & 1287K    & 425K         & 32.99             & 96.22           & 0.9510   \\
ru & Russian    & 1406K    & 877K         & 62.39             & 96.53           & 0.8413  \\
fr & French     & 979K     & 237K         & 24.20             & 96.11           & 0.9164  \\
zh & Chinese    & 1025K    & 836K         & 81.52             & 80.15           & 0.7830   \\
ar & Arabic     & 986K     & 289K         & 29.31             & 98.96           & 0.9604  \\
vi & Vietnamese & 122K     & 811K         & 66.35             & 92.46           & 0.1226  \\
ja & Japanese   & 1103K    & 768K         & 69.62             & 94.83           & 0.7297  \\
fi & Finnish    & 451K     & 270K         & 59.95             & 97.48           & 0.8303  \\
ko & Korean     & 422K     & 361K         & 85.58             & 94.76           & 0.7522  \\
tr & Turkish    & 321K     & 232K         & 72.50             & 91.76           & 0.7584  \\
ro & Romanian   & 282K     & 158K         & 56.29             & 93.38           & 0.9694  \\
cs & Czech      & 178K     & 72K          & 40.19             & 96.62           & 0.8693  \\
et & Estonian   & 195K     & 156K         & 79.74             & 99.51           & 0.8962  \\
lt & Lithuanian & 185K     & 175K         & 94.88             & 98.89           & 0.9512  \\
kk & Kazakh     & 220K     & 219K         & 99.45             & 95.54           & 0.5842  \\
lv & Latvian    & 92K      & 70K          & 76.30             & 96.35           & 0.9327  \\
hi & Hindi      & 130K     & 75K          & 57.32             & 96.66           & 0.9043  \\
ne & Nepali     & 29K      & 25K          & 84.57             & 94.74           & 0.9263  \\
my & Burmese    & 44K      & 37K          & 84.46             & 95.02           & 0.8044  \\
si & Sinhala    & 17K      & 15K          & 91.83             & 95.40           & 0.7110   \\
gu & Gujarati   & 29K      & 7K           & 24.37             & 99.60           & 0.9827  \\ \bottomrule
\end{tabular}
}
\label{tab:wd-stats}
\end{table}

\section{Article fusion}
\label{appendix:model_desc}

In order to fuse the article representations $A_\lang$ obtained by encoding the article text with mBART encoder for each language $\lang \in \{1, \dots, n\}$ into a single representation $A$, we follow the process described in \Secref{sec:method}. More formally, the $i$-th token embedding $A_i$ of $A$ is calculated as
\begin{equation*}
    \frac{1}{n} \sum\limits_{\textcolor{purple}{l}=1}^{n}\text{FF}\left(\text{LayerNorm}\left(Q_{i} + \text{softmax}\left(\frac{Q_{i}K_{\textcolor{purple}{l}}^{T}}{\sqrt{d}}\right) V_{\textcolor{purple}{l}}\right)\right),
\end{equation*}
\noindent where $Q_{i} = \left(A_q\right)_{i} W_{Q}$, $K_{\textcolor{purple}{l}} = A_{\textcolor{purple}{l}} W_{K}$,  $V_{\textcolor{purple}{l}} = A_{\textcolor{purple}{l}}  W_{V}$. Vector $\left(A_q\right)_{i}$ represents the $i$-th token embedding of $A_q$. Matrices $W_{Q}, W_{K}, W_{V}$ are trainable query, key, and value weights, and $d$ is the dimension of the key. FF corresponds to the entire feed-forward network.

\section{Error taxonomy}
\label{appendix:error taxonomy}

We adopted an iterative coding strategy by manually inspecting the data to build an understanding of the types of error that occur in the descriptions.
In each round, two authors independently labeled the same 100 randomly chosen samples, 50 where \name's \desc was preferred, and 50 where human-written description was preferred, with the possibility to expand the set of labels (error categories) in each round.
At the end of each round, the labelers discussed disagreements and the appropriateness of the selected categories, and modified them if needed.
As a stopping criterion, an average pairwise Fleiss $\kappa>0.6$ was used, and the coding process terminated after two rounds, with $\kappa=0.55$ and 0.77, respectively.

\section{Additional results}
\label{appendix:results}

In this section we present some of the examples of generated descriptions by our models. In Table~\ref{tab:summary_better},~\ref{tab:type_better}, and~\ref{tab:full_better} we present some of the examples where semantic type and existing descriptions help.

\begin{table*}[!h]
\centering
\caption{Examples where \name [no types] performs better than \name [no desc/types]}
\resizebox{0.99\textwidth}{!}{
\setlength{\tabcolsep}{3pt}
\begin{tabular}{|p{90mm}|p{55mm}|p{55mm}|p{55mm}|}
\hline
\textbf{Article}  & \textbf{Target} & \textbf{\name [no desc/types]}  & \textbf{\name [no types].}    \\ \hline
Tony LeungTony Leung Ka-Fai, nato nel 1958, chiamato anche "Big Tony" – attore cinese di Hong Kong\newline Tony Leung Chiu-Wai,   nato nel 1962, chiamato anche "Little Tony" – attore cinese di Hong Kong\newline Tony Leung Siu-Hung – attore, coreografo, regista e stuntman   cinese di Hong Kong\newline Tony Leung Hung-Wah – regista, produttore e   sceneggiatore cinese di Hong Kong & pagina di disambiguazione di un progetto Wikimedia & attore cinese                                       & pagina di disambiguazione di un progetto Wikimedia \\ \hline
The United Nations Girls' Education Initiative (UNGEI) is an   initiative launched by the United Nations in 2000 at the World Education   Forum in Dakar at the primary school Ndiarème B. It aims to reduce the gender   gap in schooling for girls and to give girls equal access to all levels of   education.    & organization    & initiative launched by   the United Nations in 2000 & organization   \\ \hline
De Golf Cup of Nations 1974 was de 3e editie van dit   voetbaltoernooi dat werd gehouden in Koeweit van 15 maart 1974 tot en met 29   maart 1974. Koeweit won het toernooi door in de finale Saoedi-Arabië te   verslaan.  & sportseizoen van een   voetbalcompetitie             & golftournooi           & sportseizoen van een   voetbalcompetitie             \\ \hline
Peter DeGraaf is a Republican member of the Kansas House of   Representatives, representing the 82nd district. He has served since May   2008.    & American politician     & Kansas House of   Representatives        & American politician   \\ \hline
The Tree of Life (Shajarat-al-Hayat) in Bahrain is a 9.75 meters   high Prosopis cineraria tree that is over 400 years old. It is on a hill in a   barren area of the Arabian Desert, 2 kilometers from Jebel Dukhan, the   highest point in Bahrain, and 40 kilometers from Manama.  & tree in Bahrain     & species of plant     & tree in Bahrain   \\ \hline
\end{tabular}
}
\label{tab:summary_better}
\end{table*}

\begin{table*}[!h]
\caption{Examples where \name [no desc] performs better than \name [no desc/types]}
\resizebox{0.99\textwidth}{!}{
\setlength{\tabcolsep}{3pt}
\begin{tabular}{|p{90mm}|p{55mm}|p{55mm}|p{55mm}|}
\hline
\textbf{Article}   & \textbf{Target}   & \textbf{\name [no desc/types]}   & \textbf{\name [no desc]}    \\ \hline
Corticotropin-releasing hormone (CRH) is a peptide hormone involved in the stress response. It is a releasing hormone that belongs to corticotropin-releasing factor family. In humans, it is encoded by the CRH gene. Its main function is the stimulation of the pituitary synthesis of   ACTH, as part of the HPA Axis.    & mammalian protein found   in Homo sapiens & chemical compound   & mammalian protein found   in Homo sapiens \\ \hline
Postcrossing is een internationaal briefkaart-uitwisselingsproject. Het werd op 13 juli 2005 opgezet door de uit Portugal afkomstige Paulo Magalhães. Postcrossing is nadien uitgegroeid tot een uitwisselingsproject met ruim 800.000 gebruikers uit 206 verschillende landen. In februari 2017 werd de 40.000.000ste postkaart die via postcrossing werd verstuurd ontvangen.   & website  & organisatie uit   Portugal & website          \\ \hline

Ligne Namboku (\begin{CJK}{UTF8}{min}南北線\end{CJK}, Nanboku-sen), littéralement ligne Sud-Nord,   est le nom donné à plusieurs lignes ferroviaires au Japon: la ligne Namboku du métro de Tokyo; la ligne Namboku du métro de Sendai; la ligne Namboku du métro de Sapporo; la ligne Namboku de la compagnie Kita-Osaka Kyuko Railway; la ligne Namboku de la compagnie Kobe Rapid Transit Railway   exploitée par la Kobe Electric Railway sous le nom de ligne Kobe Kosoku. & page d'homonymie de   Wikimedia           & ligne de chemin de fer   japonaise               & page d'homonymie de   Wikimedia           \\ \hline
The Atari Jaguar is a 64-bit home video game console developed by   Atari Corporation and designed by Flare Technology, released in North America   first on November 23, 1993. It was the sixth programmable console developed   under the Atari brand. The following list contains all of the games released   on cartridge for the Jaguar.    & Wikipedia list article    & 1993 video game console                          & Wikipedia list article    \\ \hline
The 2017 BWF World Junior Championships was the nineteenth tournament of the BWF World Junior Championships. It was held in Yogyakarta, Indonesia at the Among Rogo Sports Hall between 9 and 22 October 2017.   & badminton championships                   & 2017 edition of the   World Junior Championships & badminton championships   \\ \hline
\end{tabular}
}
\label{tab:type_better}
\end{table*}

\begin{table*}[!h]
\caption{Examples where \name performs better than other variations of \name}
\resizebox{0.99\textwidth}{!}{
\setlength{\tabcolsep}{3pt}
\begin{tabular}{|p{87mm}|p{27mm}|p{35mm}|p{33mm}|p{33mm}|p{30mm}|}
\hline
\textbf{Article}  & \textbf{Target} & \textbf{\name [no desc/types]}   & \textbf{\name [no types]}  & \textbf{\name [no desc]}   & \textbf{\name}     \\ \hline
Het Europees kampioenschap volleybal mannen 2003 vond van 5 tot en met 14 september plaats in Karlsruhe en Leipzig (Duitsland).    & sportseizoen van een volleybalcompetitie  & Volleyball-Europa    & mannenenkelspel    & mannenenkelspel    & sportseizoen van een volleybalcompetitie \\ \hline
Mohamed Shies Madhar is een Surinaams voormalig judoka. & judoka uit Suriname   & Nederlands judoka    & Nederlands judoka      & judoka    & judoka uit Suriname    \\ \hline
Guanosine monophosphate synthetase, also known as GMPS is an enzyme that converts xanthosine monophosphate to guanosine monophosphate.  & mammalian protein found   in Homo sapiens & class of enzymes     & class of enzymes     & class of enzymes  & mammalian protein found in Homo sapiens  \\ \hline
La gmina de Lubowidz est un district administratif situé en milieu rural du powiat de Żuromin dans la voïvodie de Mazovie, dans le centre-est de la Pologne.   & commune polonaise  & gmina rurale polonaise   & gmina rurale polonaise   & gmina rurale polonaise  & commune polonaise    \\ \hline
Membrane-bound transcription factor site-1 protease, or site-1 protease (S1P) for short, also known as subtilisin/kexin-isozyme 1 (SKI-1), is an enzyme that in humans is encoded by the MBTPS1 gene. S1P cleaves the endoplasmic reticulum loop of sterol regulatory element-binding protein (SREBP) transcription factors. & mammalian protein found in Homo sapiens   & protein-coding gene in   the species Homo sapiens & protein-coding gene in the species Homo sapiens & protein in Homo sapiens & mammalian protein found in Homo sapiens  \\ \hline
\end{tabular}
}
\label{tab:full_better}
\end{table*}

\end{document}
\endinput